\pdfoutput=1

\documentclass[11pt]{article}

\usepackage{ACL2023}

\usepackage{times}
\usepackage{latexsym}

\usepackage[T1]{fontenc}

\usepackage[utf8]{inputenc}

\usepackage{microtype}

\usepackage{inconsolata}
\usepackage{amsmath}
\usepackage{graphicx}
\usepackage{multirow}
\usepackage{booktabs}
\usepackage{bbding}
\usepackage{MnSymbol}
\usepackage{xcolor}
\usepackage{enumitem}
\usepackage{svg}
\usepackage{pifont}
\usepackage{lipsum}  
\usepackage[normalem]{ulem}
\usepackage{enumitem}

\newcommand\ISLTranslate{\textbf{\texttt{ISLTranslate}}}

\title{ISLTranslate: Dataset for Translating Indian Sign Language}

\author{Abhinav Joshi$^{1}$ \qquad Susmit Agrawal$^{2}$ \qquad Ashutosh Modi$^{1}$ \\
$^{1}$Indian Institute of Technology Kanpur (IIT Kanpur)\\
$^{2}$Indian Institute of Technology Hyderabad (IIT Hyderabad)\\
\texttt{\{ajoshi, ashutoshm\}@cse.iitk.ac.in},  \texttt{ai22mtech12002@iith.ac.in}
}

\begin{document}
\maketitle

\begin{abstract}
Sign languages are the primary means of communication for many hard-of-hearing people worldwide. Recently, to bridge the communication gap between the hard-of-hearing community and the rest of the population, several sign language translation datasets have been proposed to enable the development of statistical sign language translation systems. However, there is a dearth of sign language resources for the Indian sign language. This resource paper introduces \ISLTranslate, a translation dataset for continuous Indian Sign Language (ISL) consisting of 31k ISL-English sentence/phrase pairs. To the best of our knowledge, it is the largest translation dataset for continuous Indian Sign Language. We provide a detailed analysis of the dataset. To validate the performance of existing end-to-end Sign language to spoken language translation systems, we benchmark the created dataset with a transformer-based model for ISL translation. 


\end{abstract}
\section{Introduction} \label{sec:intro}

There are about 430 million hard-of-hearing people worldwide\footnote{\url{https://www.who.int/news-room/fact-sheets/detail/deafness-and-hearing-loss}} of which 63 million are in India\footnote{\url{https://nhm.gov.in/index1.php?lang=1&level=2&sublinkid=1051&lid=606}}. Sign Language is a primary mode of communication for the hard-of-hearing community. Although natural language processing techniques have shown tremendous improvements in the last five years, primarily, due to the availability of annotated resources and large language models \cite{tunstall2022natural}, languages with bodily modalities like sign languages still lack efficient language-processing systems. Recently, research in sign languages has started attracting attention in the NLP community \cite{yin-etal-2021-including,german_sign_language_koller15:cslr,turkish_sign_language_sincan2020autsl,xu-etal-2022-automatic,albanie20_bsl1k,jiang2021skeleton_realtime_sign_language,moryossef2020real_realtime_sign_language,joshi-etal-2022-cislr}. The availability of translation datasets has improved the study and development of NLP systems for sign languages like ASL (American Sign Language) \cite{american_sign_language_WASL_li2020word}, BSL (British Sign Language) \cite{BOBSL-Albanie2021bobsl}, and DGS (Deutsche Geb\"ardensprache) \cite{phoenix_dataset}. On the other hand, there is less amount of work focused on Indian Sign Language. The primary reason is the unavailability of large annotated datasets for Indian Sign Language (ISL). 
ISL being a communication medium for a large, diverse population in India, still faces the deficiency of certified translators (only 300 certified sign language interpreters in India\footnote{The statistic is as per the Indian government organization Indian Sign Language Research and Training Centre (ISLRTC): \url{http://islrtc.nic.in/}}), making the gap between spoken and sign language more prominent in India. This paper aims to bridge this gap by curating a new translation dataset for Indian Sign Language: \ISLTranslate, having $31,222$ ISL-English pairs.

\begin{figure}[t]
\centering
  \includegraphics[scale=0.06]{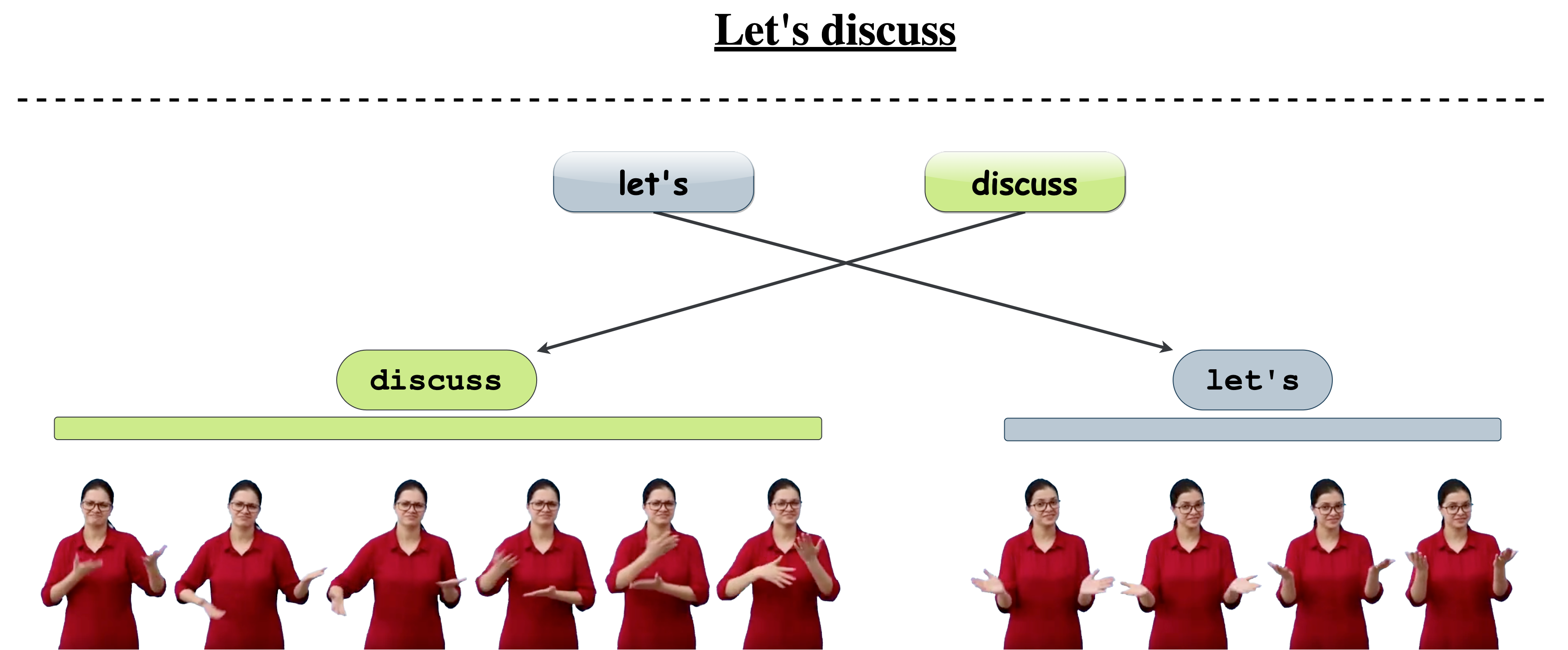}
  \caption{An example showing the translation of the phrase ``Let's discuss" in Indian Sign Language.}
  \label{fig:sign_parsing_example_lets_discuss}
  \vspace{-4mm}
\end{figure}

\noindent Due to fewer certified sign language translators for ISL, there is a dearth of educational material for the hard-of-hearing community. Many government and non-government organizations in India have recently started bridging this gap by creating standardized educational content in ISL. The created content helps build basic vocabulary for hard-of-hearing children and helps people use spoken languages to learn and teach ISL to children. Considering the standardized representations and simplicity in the vocabulary, we choose these contents for curating an ISL-English translation dataset. We choose the content specific to education material that is standardized and used across India for primary-level education. Consequently, the vocabulary used in the created content covers diverse topics (e.g., Maths, Science, English) using common daily words.  

\noindent ISL is a low-resource language, and the presence of bodily modality for communication makes it more resource hungry from the point of view of training machine learning models. Annotating sign languages at the gesture level (grouping similar gestures in different sign sentences) is challenging and not scalable. Moreover, in the past, researchers have tried translating signs into gloss representation and gloss to written language translation (Sign2Gloss2Text \cite{phoenix_dataset}). A \textit{gloss} is a text label given to a signed gesture. The presence of gloss labels for sign sentences in a dataset helps translation systems to work at a granular level of sign translation. However, generating gloss representation for a signed sentence is an additional challenge for data annotation. For \ISLTranslate, we propose the task of end-to-end ISL to English translation. Figure \ref{fig:sign_parsing_example_lets_discuss} shows an example of an ISL sign video from \ISLTranslate. The example shows a translation for the sentence ``Let's discuss'', where the signer does the sign for the word ``discuss'' by circularly moving the hands with a frown face simultaneously followed by palm lifted upwards for conveying ``let's.'' The continuity present in the sign video makes it more challenging when compared to the text-to-text translation task, as building a tokenized representation for the movement is a challenging problem. Overall, in this resource paper, we make the following contributions:
\begin{itemize}[noitemsep,nosep]
    \item We create a large ISL-English translation dataset with more than $31,222$ ISL-English pair sentences/phrases. The dataset covers a wide range of daily communication words with a vocabulary size of $11,655$. We believe making this dataset available for the NLP community will facilitate future research in sign languages with a significant societal impact. Moreover, though not attempted in this paper, we hope that \ISLTranslate\  could also be useful for sign language generation research. The dataset is made available at: \url{https://github.com/Exploration-Lab/ISLTranslate}. 
    \item We propose a baseline model for end-to-end ISL-English translation inspired by sign language transformer \cite{camgoz2020sign}. 
\end{itemize}

\section{Related Work} \label{sec:relatedwork}
In contrast to spoken natural languages, sign languages use bodily modalities, which include hand shapes and locations, head movements (like nodding/shaking), eye gazes, finger-spelling, and facial expressions. As features from hand, eye, head, and facial expressions go in parallel, it becomes richer when compared to spoken languages, where a continuous spoken sentence can be seen as a concatenated version of the sound articulated units. Moreover, translating from a continuous movement in 3 dimensions makes sign language translation more challenging and exciting from a linguistic and research perspective. 

\noindent\textbf{Sign Language Translation Datasets:} Various datasets for sign language translation have been proposed in recent years \cite{yin-etal-2021-including}. Specifically for American Sign Language (ASL), there have been some early works on creating datasets \cite{RVL-SLLL-Martinez2002PurdueRA,boston_104-dreuw07_interspeech}, where the datasets were collected in the studio by asking native signers to sign content. Other datasets have been proposed for Chinese sign language \cite{CSL_daily_dataset_Zhou2021ImprovingSL}, Korean sign language \cite{KETI_dataset_Ko2018NeuralSL}, Swiss German Sign Language - Deutschschweizer Gebardensprache (DSGS) and Flemish Sign Language - Vlaamse Gebarentaal (VGT) \cite{Content4All-open-sign-language-datasets}. In this work, we specifically target Indian Sign Language and propose a dataset with ISL videos-English translation pairs.

\begin{figure*}[t]
\centering
  \includegraphics[scale=0.05]{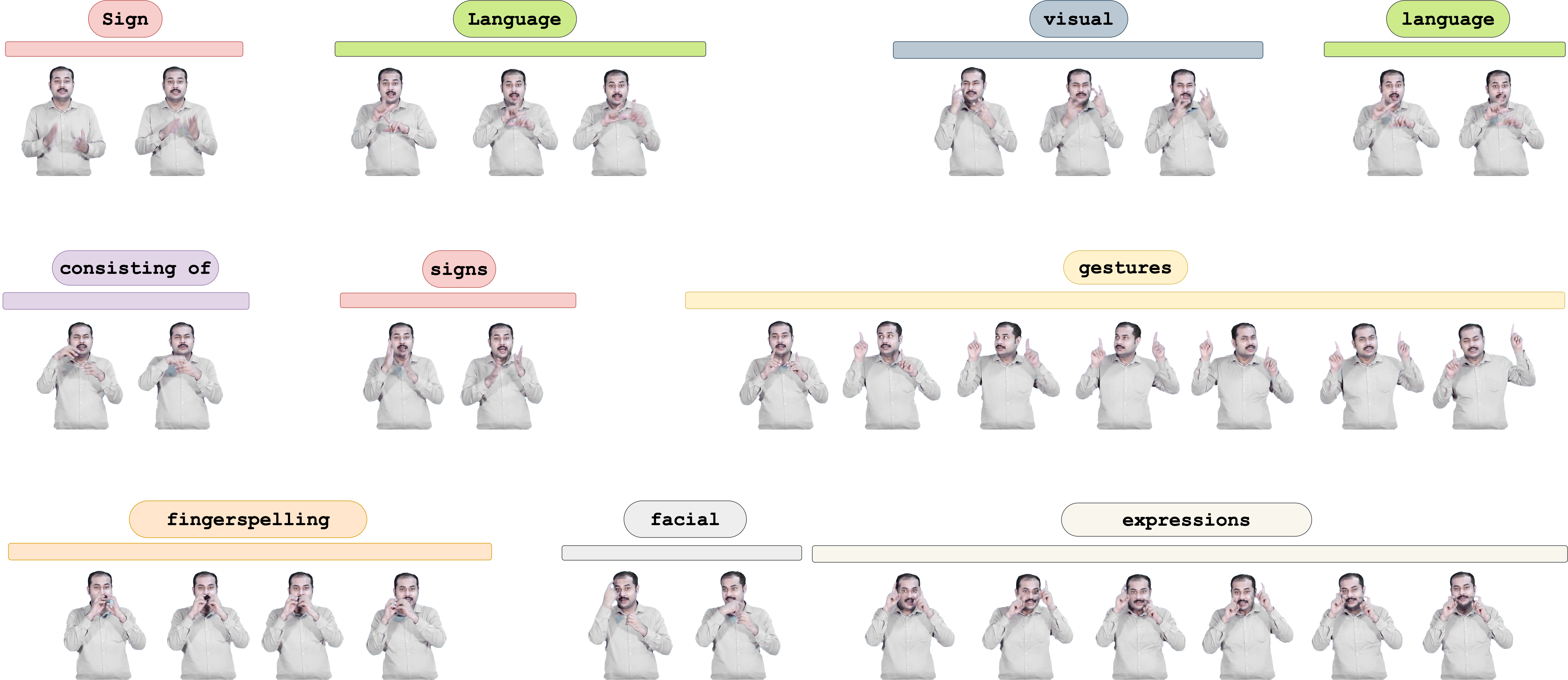}
  \caption{A sample from \ISLTranslate: ``Sign Language is a visual language consisting of signs, gestures, fingerspelling and facial expressions."}
  \label{fig:sign_parsing_example}
\end{figure*}

\begin{table}[t]
\centering
\small
\renewcommand{\arraystretch}{1}
\setlength\tabcolsep{1pt}
\begin{tabular}{lccc}
\toprule
Dataset             & Lang. & Sentences & Vocab. \\ \midrule
\begin{tabular}[l]{@{}l@{}}Purdue RVL-SLLL\\ \cite{RVL-SLLL-Martinez2002PurdueRA}\end{tabular}
& ASL      & 2.5k         & 104   \\
\begin{tabular}[l]{@{}l@{}}Boston 104\\ \cite{boston_104-dreuw07_interspeech}\end{tabular}
& ASL      & 201          & 103   \\
\begin{tabular}[l]{@{}l@{}}How2Sign\\ \cite{How2Sign_Duarte_CVPR2021}\end{tabular}
& ASL      & 35k          & 16k   \\
\begin{tabular}[l]{@{}l@{}}OpenASL\\ \cite{OpenASL}\end{tabular}
& ASL      & 98k          & 33k   \\ 

\begin{tabular}[l]{@{}l@{}}BOBSL\\ \cite{BOBSL-Albanie2021bobsl}\end{tabular}
& BSL      & 1.9k         & 78k   \\

\begin{tabular}[l]{@{}l@{}}CSL Daily\\ \cite{CSL_daily_dataset_Zhou2021ImprovingSL}\end{tabular}

& CSL      & 20.6k        & 2k    \\

\begin{tabular}[l]{@{}l@{}}Phoenix-2014T\\ \cite{phoenix_dataset}\end{tabular}
& DGS      & 8.2          & 3K    \\

\begin{tabular}[l]{@{}l@{}}SWISSTXT-Weather\\ \cite{Content4All-open-sign-language-datasets}\end{tabular}
& DSGS     & 811          & 1k    \\

\begin{tabular}[l]{@{}l@{}}SWISSTXT-News\\ \cite{Content4All-open-sign-language-datasets}\end{tabular}
& DSGS     & 6k           & 10k   \\

\begin{tabular}[l]{@{}l@{}}KETI\\ \cite{KETI_dataset_Ko2018NeuralSL}\end{tabular}
& KSL      & 14.6k        & 419   \\

\begin{tabular}[l]{@{}l@{}}VRT-News\\ \cite{Content4All-open-sign-language-datasets}\end{tabular}
& VGT      & 7.1k         & 7k    \\
\midrule
 \begin{tabular}[l]{@{}l@{}}ISL-CSLRT \\ 
 \cite{resource_indian_eelakkiya2021isl_ISL_CSLRT}
\end{tabular}
& ISL      & 100          &     -  \\
\ISLTranslate\  (ours) & ISL      & 31k          & 11k    \\ \bottomrule
\end{tabular}
\caption{Comparison of continuous sign language translation datasets.}
\vspace{-4mm}
\label{tab:data-comparison}
\end{table}

\noindent\textbf{End-to-End Sign Language Translation Systems:} Most of the existing approaches for sign language translation \cite{phoenix_dataset, info13050220, De_Coster_2021_AT4SSL} depend on intermediate gloss labels for translations. 
As glosses are aligned to video segments, they provide fine one-to-one mapping that facilitates supervised learning in learning effective video representations. Previous work \cite{phoenix_dataset} has reported a drop of about $10.0$ in BLEU-4 scores without gloss labels. However, considering the annotation cost of gloss-level annotations, it becomes imperative to consider gloss-free sing language translation approaches. Moreover, the gloss mapping in continuous sign language might remove the grammatical aspects from the sign language.
Other recent works on Sign language translation include \citet{Voskou2021StochasticTN, yin-read-2020-better}, which try to remove the requirement for a glossing sequence for training and proposes a transformer-based architecture for end-to-end translations. We also follow a gloss-free approach for ISL translation.

\section{\ISLTranslate\ } \label{sec:dataset}

\begin{table*}[t]
\centering
\small
\renewcommand{\arraystretch}{1}
\setlength\tabcolsep{1pt}
\begin{tabular}{ll}
\toprule
\textbf{\ISLTranslate\ Translations}                                   & \textbf{ISL-Signer Translations (references)}                          \\ \midrule
Birbal started smiling. When it was his turn, he went near the line. & Birbal started smiling. He \textcolor{blue}{turned towards the drawn line.}              \\
Discuss with your partner what Birbal would do.                      & Discuss with your partner what Birbal would do.                        \\
\textcolor{red}{Birbal drew a longer line.}                                           & \textcolor{blue}{Under the drawn line, Birbal}                                           \\
\textcolor{red}{under the first one.}                                                 & \textcolor{blue}{drew a longer line.}                                                    \\
saw what he drew and said.                                           & \textcolor{blue}{and wondered}                                                           \\
That's true, the first line is shorter now.                          & That's true, the first line is shorter now.                            \\
One day, Akbar drew a line \textcolor{red}{on the floor} and ordered.                 & One day, Akbar drew a line and ordered                                 \\
Make this line shorter.                                              & Make this line shorter.                                                \\
Rita is shorter than Radha.                                          & Rita is \textcolor{blue}{short and the other is} Radha.                                  \\
Rajat is taller than Raj.                                            & Rajat is taller \textcolor{blue}{and the other} is Raj.                                  \\
but don't rub out any part of it.                                    & but don't rub out any part of it.                                      \\
\textcolor{red}{Try to draw Rajat's taller than Raj.}                                 & \textcolor{blue}{First draw Rajat as taller, then draw Raj on the right.}                \\
No one knew what to do.                                              & No one knew what to do.                                                \\
Each minister looked at the line and was puzzled.                    & Each minister looked at the line and was puzzled.                      \\
No one could think of any way to make it longer.                     & No one could think of any way to make it longer.                       \\
\textcolor{red}{Have you seen the fine wood carving?}                                 & \textcolor{blue}{Look at its architecture.}                                              \\
Most houses had a separate                                           & Most houses had a separate \textcolor{blue}{bathing}                                     \\
\textcolor{red}{Beding area.}                                                         & \textcolor{blue}{separate}                                                               \\
and some had wells to supply water.                                  & and some had wells to supply water.                                    \\
Many of these cities had covered drains.                             & Many of these cities had covered drains.                               \\
Notice how carefully these were laid out in straight lines.          & Notice how carefully these were laid out in straight lines.           \\ \bottomrule
\end{tabular}
\caption{The Table shows a sample of English translations present in \ISLTranslate\ compared to sentences translated by ISL Signer for the respective ISL videos. The exact ISL-Signer Translations were used as reference sentences for computing translation metric scores reported in Table \ref{tab:islrtc-reference-translations-metrics}. \textcolor{blue}{Blue} and \textcolor{red}{Red} colored text highlight the difference between semi-automatically generated English sentences and gold sentences generated by the ISL instructor.}
\label{tab:islrtc-reference-sample-sentences}
\vspace{-2mm}
\end{table*}

\begin{table}[t]
\centering
\small
\renewcommand{\arraystretch}{1}
\setlength\tabcolsep{5pt}
\begin{tabular}{cc}
\toprule
\textbf{Metric} & \textbf{Score} \\
\midrule
BLEU-1          & 60.65         \\
BLEU-2          & 55.07         \\
BLEU-3          & 51.43         \\
BLEU-4          & 48.93         \\
METEOR          & 57.33         \\
WER             & 61.88         \\
ROUGE-L         & 60.44        \\
\bottomrule
\end{tabular}
\caption{The Table shows the translation scores for a random sample of 291 pairs from \ISLTranslate\  when compared to references translated by the ISL instructor.}
\label{tab:islrtc-reference-translations-metrics}
\vspace{-4mm}
\end{table}

\ISLTranslate\ is a dataset created from publicly available educational videos produced by the ISLRTC organization and made available over YouTube. These videos were created to provide school-level education to hard-of-hearing children. The videos cover the NCERT\footnote{ \url{https://en.wikipedia.org/wiki/National\_Council\_of\_Educational\_Research\_and\_Training}} standardized English educational content in ISL. As the targeted viewers for these videos are school children and parents, the range of words covered in the videos is beginner-level. Hence, it provides a good platform for building communication skills in ISL. The videos cover various NCERT educational books for subjects like science, social science, and literature. A single video (about 15-30 minutes) usually covers one chapter of a book and simultaneously provides an audio voice-over (in English) conveying the same content. Apart from ISLRTC's educational sign videos which make up a significant portion of \ISLTranslate, we also use another resource from Deaf Enabled Foundations (DEF) (\url{https://def.org.in/}). DEF videos consist of words with respective descriptions and example sentences for the same words, along with the text transcriptions available in the descriptions\footnote{Example video: \url{https://www.youtube.com/watch?v=429wv1kvK_c}}. We split the DEF Sign videos into multiple segments using visual heuristics for separating segments corresponding to words, descriptions, and examples. In total, \ISLTranslate\  consists of $2685$ videos ($8.6\%$) from DEF, and the remaining $28537$ videos ($91.4\%$) are from ISLRTC.

\noindent\textbf{\ISLTranslate\ Creation:} We use the audio voice-over (by analyzing the speech and silence parts) to split the videos into multiple segments. Further, these segments are passed through the SOTA speech-to-text model \cite{Whisper} to generate the corresponding text. As the generated text is the same as present in the book chapters' text, verifying the generated sample was easy and was done by manually matching them with the textbook. In general, we found automatically transcribed text to be of high quality; nevertheless, incorrectly generated text was manually corrected with the help of content in the books. 

\noindent Figure \ref{fig:sign_parsing_example} shows an example (from \ISLTranslate) of a long sentence and its translation in ISL. The frames in the figure are grouped into English words and depict the continuous communication in ISL. Notice the similar words in the sentence, ``sign/signs'' and ``language.'' (also see a visual representation of Sign Language\footnote{\url{https://www.youtube.com/watch?v=SInKhy-06qA}}). As common to other natural languages, representations (characters/gestures) of different lengths are required for communicating different words. In \ISLTranslate, we restrict to the sentence/phrase level translations. The dataset is divided into train, validation, and test splits (Details in App. \ref{app:data-split}).  App. Figure \ref{fig:token_distribution} shows the distribution of the number of samples in various splits. 

\noindent\textbf{Comparison with Other Continuous Sign-Language Datasets:}  We primarily compare with video-based datasets containing paired continuous signing videos and the corresponding translation in written languages in Table \ref{tab:data-comparison}. To the best of our knowledge, we propose the largest dataset for ISL.  

\noindent\textbf{Data Cleaning and Preprocessing:} The videos (e.g., App. Fig. \ref{fig:sample-video}) contain the pictures corresponding book pages. We crop the signer out of the video by considering the face location as the reference point and removing the remaining background in the videos.



\noindent\textbf{Noise Removal in \ISLTranslate:} 
As the \ISLTranslate\  consists of videos clipped from a longer video using pauses in the available audio signal, there are multiple ways in which the noises in the dataset might creep in. While translating the text in the audio, a Signer may use different signs that may not be the word-to-word translation of the respective English sentence. Moreover, though the audio in the background is aligned with the corresponding signs in the video, it could happen in a few cases that the audio was fast compared to the corresponding sign representation and may miss a few words at the beginning or the end of the sentence. We also found a few cases where while narrating a story, the person in the audio takes the character role by modifying speech to sound like the designated character speaking the sentence. For example, in a story where a mouse is talking, instead of saying the sentence followed by the ``said the mouse'' statement, the speakers may change their voice and increase the pitch to simulate dialogue spoken by the mouse. In contrast, in the sign language video, a person may or may not take the role of the mouse while translating the sentence to ISL. 


\noindent\textbf{\ISLTranslate\ Validation:} To verify the reliability of the sentence/phrase ISL-English pairs present in the dataset, we take the help of a certified ISL signer. Due to the limited availability of certified ISL Signers, we could only use a small randomly selected sign-text pairs sample ($291$ pairs) for human translation and validation. We ask an ISL instructor to translate the videos (details in App. \ref{app:annotation}). Each video is provided with one reference translation by the signers. Table \ref{tab:islrtc-reference-sample-sentences} shows a sample of sentences created by the ISL instructor. To quantitatively estimate the reliability of the translations in the dataset, we compare the English translation text present in the dataset with the ones provided by the ISL instructor. Table \ref{tab:islrtc-reference-translations-metrics} shows the translation scores for $291$ sentences in \ISLTranslate. Overall, the BLEU-4 score is $48.94$, ROUGE-L \cite{lin-2004-rouge} is $60.44$, and WER (Word Error Rate) is $61.88$. To provide a reference for comparison, for text-to-text translations BLEU score of human translations ranges from 30-50 (as reported by \citet{papineni-etal-2002-bleu}, on a test corpus of about 500 sentences from 40 general news stories, a human translator scored 34.68 against four references). We speculate high reliability over the translations present in the \ISLTranslate\  with a BLEU score of 48.93 compared against the reference translations provided by certified ISL Signer. Ideally, it would be better to have multiple reference translations available for the same signed sentence in a video; however, the high annotation effort along with the lower availability of certified ISL signers makes it a challenging task.


\section{Baseline Results} \label{sec:all-results}

\noindent Given a sign video for a sentence, the task of sign language translation is to translate it into a spoken language sentence (English in our case). For benchmarking \ISLTranslate, we create a baseline architecture for ISL-to-English translation. We propose an ISL-pose to English translation baseline (referred to as Pose-SLT) inspired by Sign Language Transformer (SLT) \cite{camgoz2020sign}. Sign language transformer uses image features with transformers for generating text translations from a signed video. However, considering the faster real-time inference of pose estimation models \cite{selvaraj-etal-2022-openhands}, we use pose instead of images as input. We use the Mediapipe pose estimation pipeline\footnote{\url{https://ai.googleblog.com/2020/12/mediapipe-holistic-simultaneous-face.html}}. A similar SLT-based pose-to-Text approach was used by \citet{saunders2020progressivetransformers}, which proposes Progressive Transformers for End-to-End Sign Language Production and uses SLT-based pose-to-text for validating the generated key points via back translation (generated pose key points to text translations). Though the pose-based approaches are faster to process, they often perform less than the image-based methods. For the choice of holistic key points, we follow \citet{selvaraj-etal-2022-openhands}, which returns the 3D coordinates of 75 key points (excluding the face mesh). Further, we normalize every frame’s key points by placing the midpoint of shoulder key points to the center and scaling the key points using the distance between the nose key point and the shoulders midpoint. We use standard BLEU and ROUGE scores to evaluate the obtained English translations (model hyperparameter details in  App.\ref{app:hyper}). Table \ref{tab:results} shows the results obtained for the proposed architecture. Poor BLEU-4 result highlights the challenging nature of the ISL translation task. The results motivate incorporating ISL linguistic priors into data-driven models to develop better sign language translation systems.

\begin{table}[t]
\centering
\small
\renewcommand{\arraystretch}{1}
\setlength\tabcolsep{1pt}
\begin{tabular}{cccccc}
\toprule
Model         & BLEU-1 & BLEU-2 & BLEU-3 & BLEU-4 & ROUGE-L   \\ \midrule
Pose-SLT & 13.18 & 8.77 & 7.04 & 6.09 & 12.91 \\ \bottomrule
\end{tabular}
\caption{The table shows translation scores obtained for  the baseline model.}
\vspace{-4mm}
\label{tab:results}
\end{table}

\section{Conclusion}\label{sec:conclusion}

We propose \ISLTranslate, a dataset of 31k ISL-English pairs for ISL. We provide a detailed insight into the proposed dataset and benchmark them using a sign language transformer-based ISL-pose-to-English architecture. Our experiments highlight the poor performance of the baseline model, pointing towards a significant scope for improvement for end-to-end Indian sign language translation systems. We hope that \ISLTranslate\ will create excitement in the sign language research community and have a significant societal impact. 

\section*{Limitations}\label{sec:limitation}

This resource paper proposes a new dataset and experiments with a baseline model only. We do not focus on creating new models and architectures. In the future, we plan to create models that perform much better on the \ISLTranslate\  dataset. Moreover, the dataset has only 31K video-sentence pairs, and we plan to extend this to enable more reliable data-driven model development. In the future, we would also like to incorporate ISL linguistic knowledge in data-driven  models. 

\section*{Ethical Concerns}

We create a dataset from publicly available resources without violating copyright. We are not aware of any ethical concerns regarding our dataset. Moreover, the dataset involves people of Indian origin and is created mainly for Indian Sign Language translation. The annotator involved in the dataset validation is a hard-of-hearing person and an ISL instructor, and they performed the validation voluntarily.

\section*{Acknowledgements} 
We want to thank anonymous reviewers for their insightful comments. We want to thank Dr. Andesha Mangla (\url{https://islrtc.nic.in/dr-andesha-mangla}) for helping in translating and validating a subset of the \ISLTranslate\  dataset. 
\bibliographystyle{acl_natbib}
\bibliography{references}

\clearpage
\newpage
\appendix
\section*{Appendix}

\section{Data Splits} \label{app:data-split}

Data splits for \ISLTranslate\  are shown in Table \ref{tab:train-val-test-split}. 

\begin{table}[h]
\centering
\small
\renewcommand{\arraystretch}{1}
\begin{tabular}{cccc}
\toprule
          & Train & Validation & Test \\ \midrule
\# Pairs & 24978 (80\%) & 3122 (10\%)   & 3122 (10\%) \\ 
\bottomrule
\end{tabular}
\caption{The table shows the train, validation, split for the  \ISLTranslate.}
\label{tab:train-val-test-split}
\end{table}

\section{ISLTranslate Word Distribution}

\begin{figure}[h]
\centering
  \includegraphics[width=\linewidth]{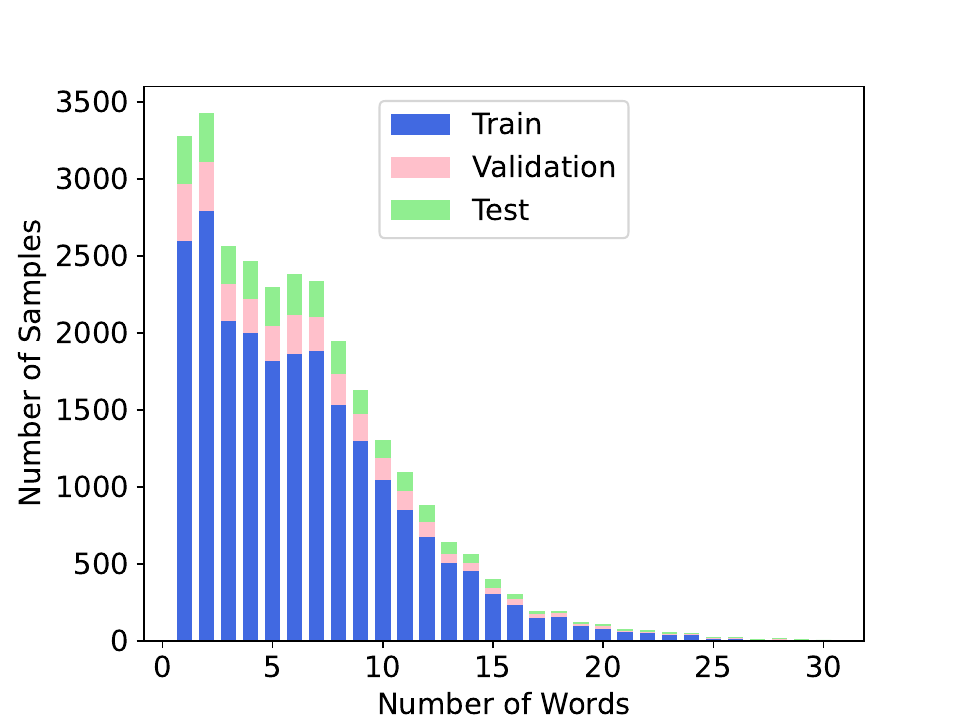}
  \caption{Distribution of the number of samples in the train, validation, and test splits of \ISLTranslate.}
  \label{fig:token_distribution}
  \vspace{-3mm}
\end{figure}

\section{Annotation Details} \label{app:annotation}
We asked a certified ISL instructor to translate and validate a random subset from the dataset. The instructor is a hard-of-hearing person and uses ISL for communication; hence they are aware of the subtitles of ISL. Moreover, the instructor is an assistant professor of sign language linguistics. The instructor is employed with ISLRTC, the organization involved in creating the sign language content; however, the instructor did not participate in videos in \ISLTranslate. The instructor performed the validation voluntarily. It took the instructor about 3 hours to validate 100 sentences. They generated the English translations by looking at the video. 

\begin{figure}[h!]
\centering
  \includegraphics[width=0.9\linewidth]{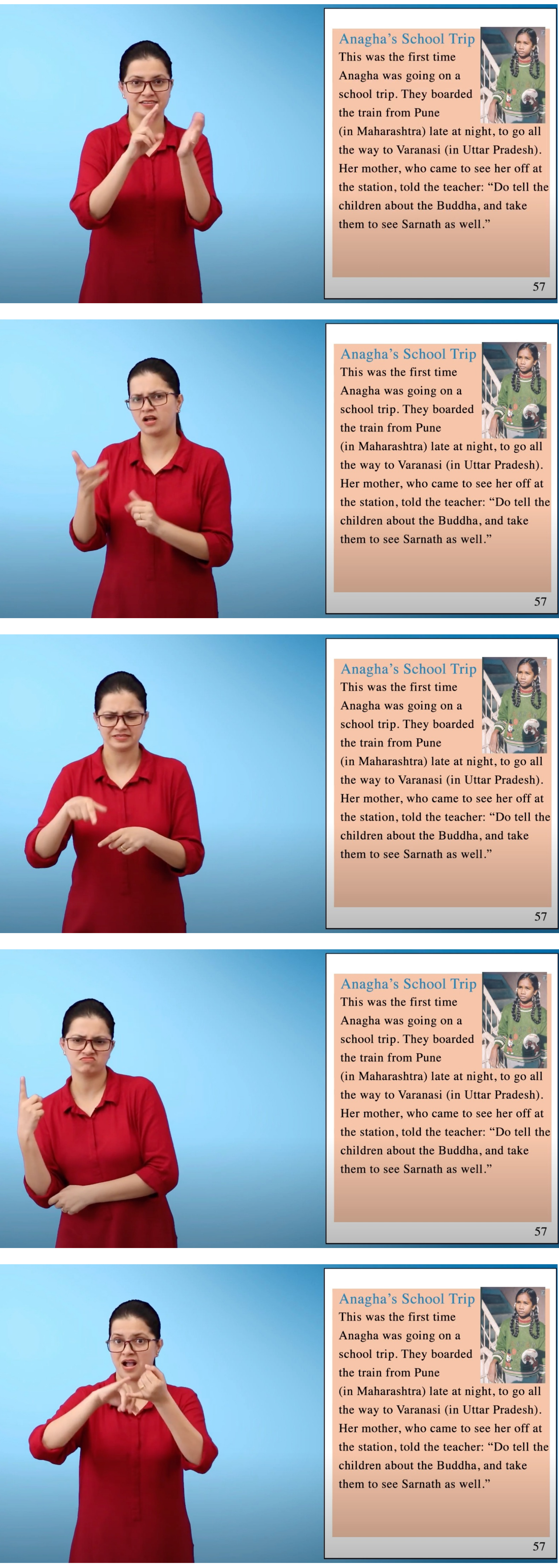}
  \caption{The figure shows an example of the educational content video where the signer signs for the corresponding textbook.}
  \label{fig:sample-video}
  \vspace{-4mm}
\end{figure}

\section{Hyperparameters and Training}\label{app:hyper}
We follow the code base of SLT \cite{camgoz2020sign} to train and develop the proposed SLT-based pose-to-text architecture by modifying the input features to be sign-pose sequences generated by the mediapipe. The model architecture is a transformer-based encoder-decoder consisting of $3$ transformer layers each for both encoder and decoder. 
We use the Adam optimizer \cite{kingma2014adam} 
with a learning rate of 0.0001, $\beta = (0.9, 0.999)$ and weight decay of  $0.0001$
for training the proposed baseline with a batch size of 32.

\end{document}